\documentclass{article}
\usepackage{spconf,amsmath,graphicx}
\usepackage{booktabs}
\usepackage{comment}
\usepackage{amssymb}
\usepackage{multirow}
\usepackage{url}
\usepackage{hyperref}

\def\L{{\cal L}}

\title{IMPROVING TRANSFORMER-BASED SPEECH RECOGNITION USING UNSUPERVISED PRE-TRAINING}
\name{Dongwei Jiang, Xiaoning Lei, Wubo Li, Ne Luo, Yuxuan Hu, Wei Zou, Xiangang Li}
\address{
  AI Labs, Didi Chuxing, Beijing, China \\
  \{jiangdongwei, leixiaoning, liwubo, luone, huyuxuan\_i, zouwei, lixiangang\}@didiglobal.com
  }

\begin{document}
\ninept
\maketitle
\begin{abstract}
	Speech recognition technologies are gaining enormous popularity in various industrial applications. However, building a good speech
	 recognition system usually requires large amounts of transcribed data, which is expensive to collect. To tackle
	  this problem, an unsupervised pre-training method called Masked Predictive Coding is proposed, which
	   can be applied for unsupervised pre-training with Transformer based model. Experiments on HKUST
		show that using the same training data, we can achieve CER 23.3\%,  exceeding the best end-to-end model by over 0.2\% absolute CER. 
		With more pre-training data, we can further reduce the CER to 21.0\%, or
		 a 11.8\% relative CER reduction over baseline.
\end{abstract}
\section{INTRODUCTION}
\label{sec:intro}
	Current industrial end-to-end automatic speech recognition (ASR) systems rely heavily on large amount of high quality
	 transcribed audio data. However, transcribed data take substantial effort to obtain in industrial applications,
	  while at the same time a lot of un-transcribed data exist in online systems and cost little to collect. It is worthwhile
	   to explore how to effectively use un-transcribed data to improve the performance of speech recognition systems
	    when labeled data are limited. 

	Recently, unsupervised pre-training has shown promising results in several areas, including Computer Vision (CV)
	 \cite{DBLP:conf/iccv/DoerschGE15}, Natural Language Processing (NLP) \cite{DBLP:conf/naacl/DevlinCLT19, DBLP:conf/naacl/EdunovBA19}
	  and so on \cite{radford2018improving, Schneider_2019, lian2018improving}. One work that stands out among these unsupervised
	   pre-training methods is Bidirectional Encoder Representations from Transformers (BERT)
		\cite{DBLP:conf/naacl/DevlinCLT19}, which used a Masked Language Model (MLM) pre-training objective and 
		obtained new state-of-the-art results on eleven NLP benchmarks.
		 
	In speech area, researchers also proposed some unsupervised pre-training algorithms. Contrastive Predictive Coding (CPC)
	 \cite{oord2018representation} is one of those unsupervised approaches that extract representation from data by predicting
	  future information. There are multiple concurrent approaches that generalize this approach and applied it in learning
	   speaker representation \cite{Ravanelli_2019}, extracting speech representation \cite{Schneider_2019, Pascual_2019} 
	   and performing speech emotion recognition \cite{lian2018improving}. Apart from CPC, \cite{Chung_2019} proposed a different
		pre-training model called Autoregressive Predictive Coding (APC) and also got comparable results on phoneme
		 classification and speaker verification. 
	
	Both CPC and APC belong to the family of future predictive coding. One constraint of these methods is they can only
	 be applied in uni-directional models but not Transformer based models. Originally proposed as a better replacement 
	 of Recurrent Neural Network (RNN) for neural machine translation, Transformer based models got a lot of traction in
	  the field of speech recognition recently \cite{DBLP:conf/icassp/DongWX19, karita2019comparative, dong2018speech, ZhouDXX18}.
	   The benefits of Transformer based models include faster training speed, better utilization of related context information
	    and better performance over RNN in many speech recognition benchmarks \cite{karita2019comparative}. In this work, 
	 we get intuition from BERT and propose a simple and effective pre-training method called Masked Predictive Coding (MPC). The design of MPC 
	  bears a lot of similarity to the Masked Language Model (MLM) objective proposed in BERT. It enables unsupervised pre-training 
	  for models that incorporate context from both directions.

	To test the effectiveness MPC,  experiments were conducted on HKUST and AISHELL-1 with different pre-training data.
	Experimental results on HKUST
		show that using the same training data, we can achieve CER 23.3\%,  exceeding the best end-to-end model by over 0.2\% absolute CER.
	 Using about 1500 hours of open source Mandarin data for unsupervised pre-training, we
	 can reduce CER of a strong Transformer based baseline by up to 3.8\% while on AISHELL-1, pre-training with the same data 
	 reduces CER by 14.7\%. To better understand the effect of pre-training data size and speaking style on 
	  fine-tuning task, experiments were also conducted with our internal dataset. With matching pre-training dataset and 
	  more data, our best performing HKUST model achieved a CER reduction of 11.8\% while our best performing AISHELL-1
	   model achieved a CER reduction of 22.1\% over baseline.

	%The rest of the paper is organized as follows: Section 2 reviews related work to MPC. Section 3 explains MPC
	% and our training procedure in detail. Experimental setup and results are presented in Section 4. Section 5 concludes the paper with discussion 
	% and future work.

% Among these one of the most commonly used unsupervised learning techniques is unsupervised pre-training of neural networks, either being feature-based or fine-tuning based \cite{DBLP:conf/naacl/DevlinCLT19}. The insight behind this is to learn representations towards better initialization than the standard random initialization, and capture more intricate dependencies between parameters, giving neural networks better generalization, regularization and robustness \cite{DBLP:journals/jmlr/ErhanMBBV09, DBLP:journals/jmlr/GlorotB10, DBLP:journals/jmlr/ErhanBCMVB10}.
% Predicting future or neighboring is an effective strategy for unsupervised representation learning, and it has been used to learn representations for images and words \cite{DBLP:conf/iccv/DoerschGE15, DBLP:conf/naacl/DevlinCLT19, DBLP:journals/corr/abs-1301-3781}. 

\section{RELATED WORK}
\label{sec:related_work}
	
	% cpc
	Contrastive Predictive Coding (CPC) is proposed by van den Oord et al. \cite{oord2018representation} to learn representations
	 from high-dimensional signal in an unsupervised manner by utilizing next step prediction. The main building blocks of CPC include a
		non-linear encoder $ g_{enc} $ and an autoregressive model $ g_{ar} $.
		 Given an input sequence $ (x_1, x_2, \dots, x_T) $, $ g_{enc} $ encodes observations $ x_t $ to a latent embedding space $ z_t = g_{enc}(x_t) $
		  while $ z_t $ is then fed to $ g_{ar} $ to produce a context representation $ c_t = g_{ar}(z_{\leq t}) $. Targeting at predicting future
		   observations $ x_{t + k} $, a density ratio $ f(x_{t + k}, c_t) $ is modelled to maximally preserve the mutual information between $ x_{t + k} $
			and $ c_t $. This design releases the model from modelling complex relationship in the data which might be unnecessary for extracting a
			 good representation towards downstream tasks. 

	To optimize the encoder $ g_{enc} $ and the autoregressive model $ g_{ar} $, the contrastive loss is minimized:
	\begin{equation}
		\L_N = -\displaystyle \mathop{\mathbb{E}}_{X}[log\frac{f(x_{t + k}, c_t)}{\sum_{x_j\in X}f_k(x_j, c_t)}],
	\end{equation}
	 where $ N $ represents number of samples in $ X = \{x_1, x_2, \dots, x_N \} $, with one positive sample from
	 distribution $ p(x_{t + k} | c_t) $ and the rest being negative samples from distribution $ p(x_{t + k}) $. 

	% apc
	Though also belonging to the family of predictive models, APC got intuition from recent progress in NLP based on language model
	 pre-training \cite{radford2018improving, radford2019language, PetersNIGCLZ18} and chose to directly optimize L1 loss between
	  input sequence and output sequence. Specifically, given input sequence $ (x_1, x_2, \dots, x_T) $ and output sequence
	   $ (y_1, y_2, \dots, y_T) $, the loss is defined as:
	\begin{equation}
		\L = \sum_{i = 1}^{T - n}|x_{i + n} - y_i|.
	\end{equation}

\section{PROPOSED METHOD}
\label{sec:proposed_method}

\begin{figure}[h]
		\begin{minipage}[b]{1.0\linewidth}
		\centering
		\centerline{\includegraphics[clip, trim=3cm 1.5cm 3.5cm 4cm, width=0.8\textwidth]{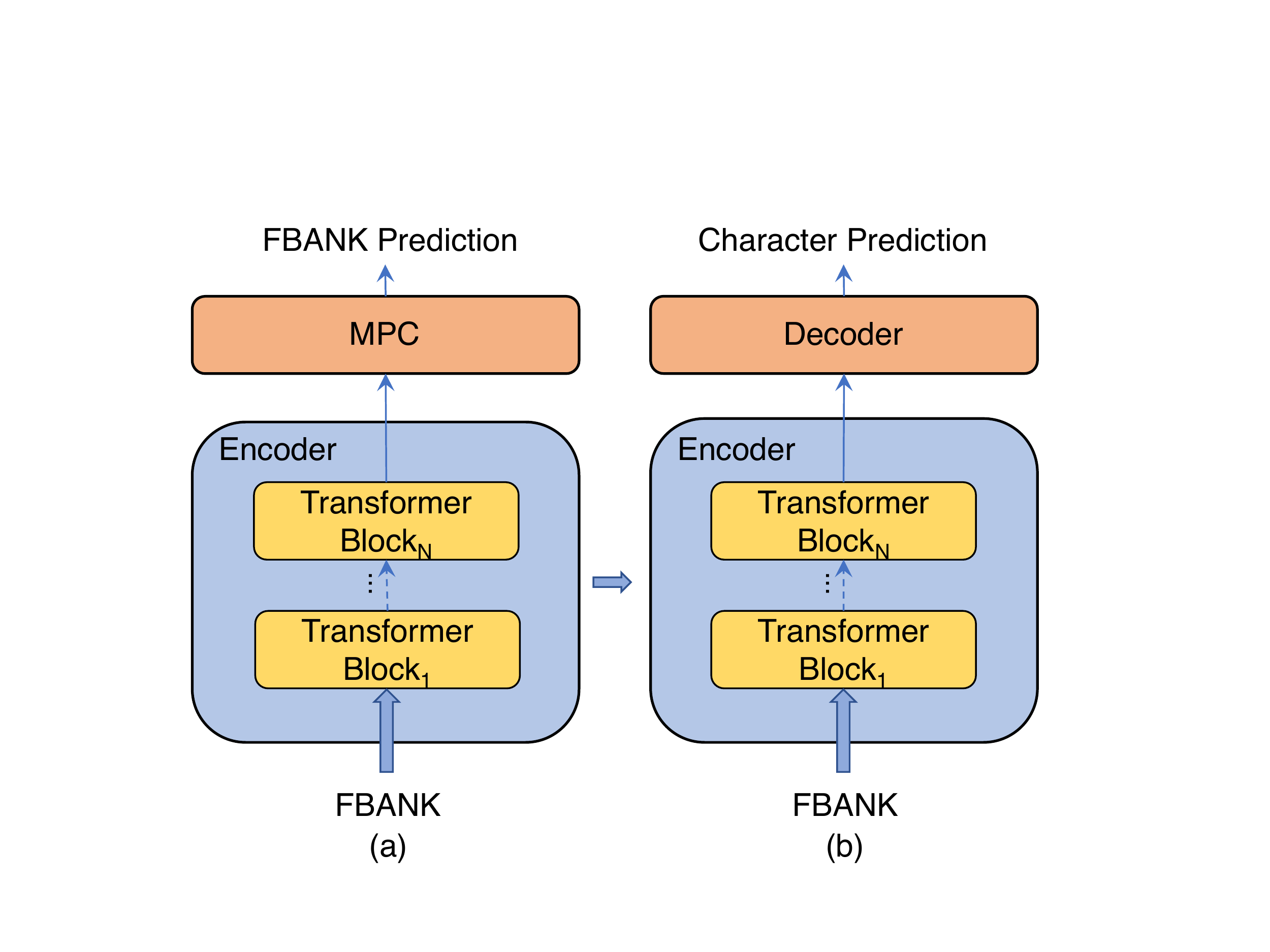}}
		\end{minipage}
		
		\caption{Our training procedure: (a) Pre-training: The encoder is used to predict FBANK utilizing Masked Predictive Coding.
		(b) Fine-tuning: Transformer decoder was added after the encoder, then the model is fine-tuned for character prediction.}
		\label{fig:Structure}
	\end{figure}

	The overall training procedure is shown in Fig. \ref{fig:Structure}. Our training consists of two stages, pre-training on unsupervised 
	data and fine-tuning on supervised data. To introduce minimum changes to the model structure, predictive coding was directly performed on
 	 FBANK input and encoder output. The encoder output is projected to the same dimension as FBANK input in all of our experiments.
  	 After the unsupervised pre-training procedure, we remove added layer for predictive coding and plug in Transformer decoder for fine-tuning
	 on downstream ASR tasks. Unlike previous work with predictive coding \cite{Schneider_2019}, our setup does not introduce any
	  additional parameters into speech recognition model. All model parameters are end-to-end trainable in the fine-tuning stage.

\begin{figure}[h]

		\begin{minipage}[b]{1.0\linewidth}
			\centering
			\centerline{\includegraphics[clip, trim=4cm 3.7cm 4.7cm 3.3cm, width=0.97\textwidth]{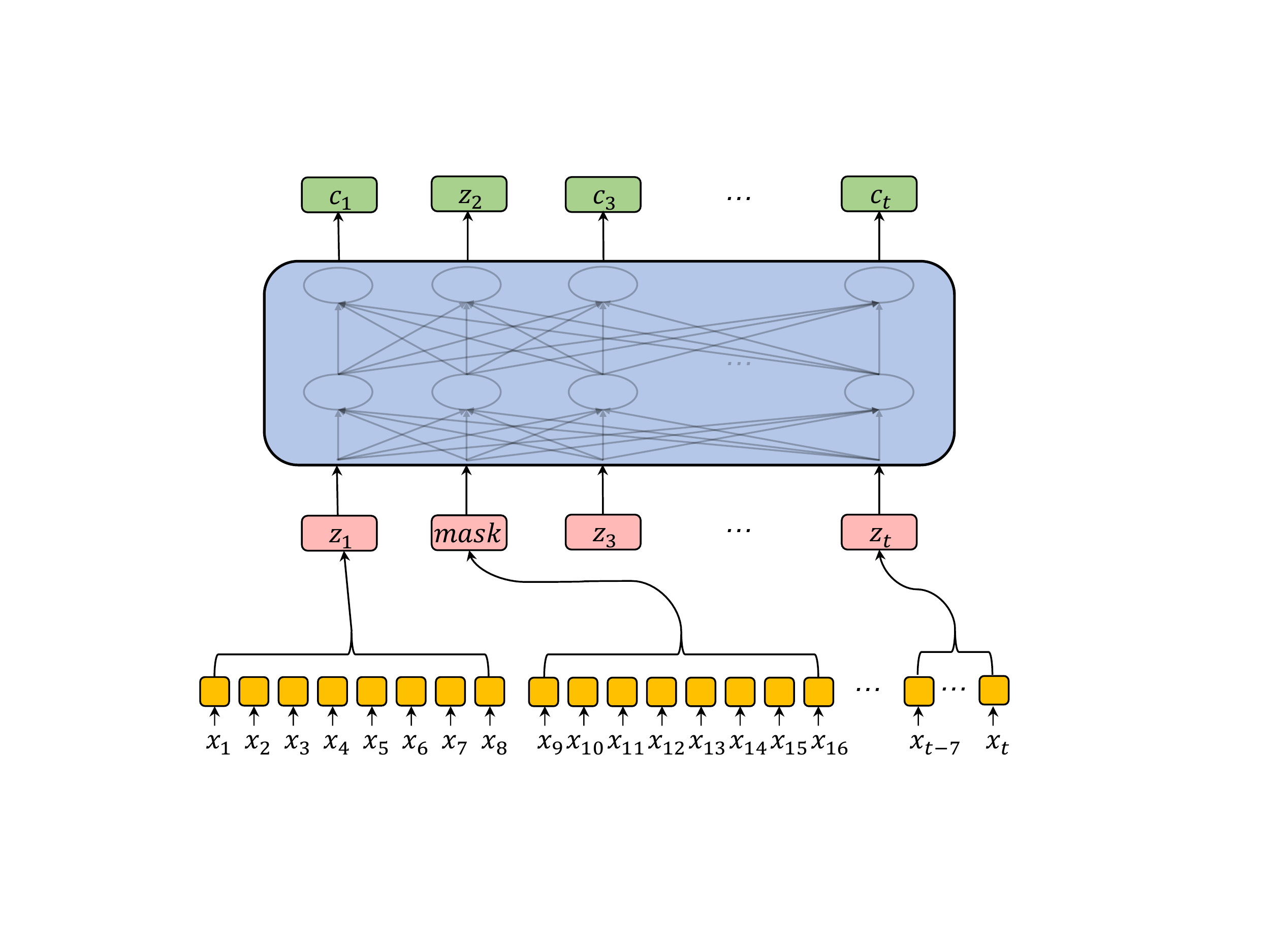}}
		\end{minipage}
	
		\caption{Masked Predictive Coding with eight-fold downsample.}
		\label{fig:MPC}
		
	\end{figure}

	% mpc
	 Inspired by BERT \cite{DBLP:conf/naacl/DevlinCLT19}, MPC uses Masked-LM (MLM) like structure to perform predictive coding on 
	 Transformer based models. 
	 The overall structure of MPC is depicted in Fig. \ref{fig:MPC}. Similar setup to \cite{DBLP:conf/naacl/DevlinCLT19} was used, 
	 where 15\% of the tokens in each sequence are
	 chosen to be masked during pre-training procedure, which in our case are frames in each audio. The chosen
	 frames are replaced with zero vectors for 80\% of the time, with frames from random positions 10\% of the time, and kept 
	 to be the same in the rest of the time. L1 loss is computed between masked input FBANK features and encoder output at
	 corresponding position. Dynamic masking proposed in \cite{liu2019roberta} was also adopted
	 where the masking pattern is generated every time a sequence is fed to the model.

	 One unique characteristic of sequence-to-sequence with attention ASR models is it usually applies downsampling in the encoder \cite{ChanJLV16}.
	 Previous research shows temporal pooling encourages effective encoding in different temporal resolution and makes alignments
	  in the decoding easier \cite{ChanJLV16}. Like CPC and APC, we first tried not using downsampling at all in pre-training stage but didn't get 
	  much improvements over the baseline. We suspect the local smoothness of speech signals combined with the
	   powerful modelling ability of Transformer make convergence too easy in this setup. As a result, the learned parameters contain very little information.
		So in this work, downsampling was applied on input feature before feeding in encoder in pre-training stage. Downsampling in fine-tuning stage
		 is applied inside the model.

\section{EXPERIMENTS}
\label{sec:experiments}

	\subsection{Data}
	\label{sec:data}
		With reproducibility in mind, some open source Mandarin datasets were collected from Linguistic Data Consortium (LDC) and OpenSLR for pre-training. 
		The corpora we collected include: HKUST Mandarin Telephone Speech Corpus (HKUST/MTS) \cite{DBLP:conf/iscslp/LiuFYCHG06},
		\begin{table*}[ht]
			\centering
			\caption{Character Error Rates on HKUST and AISHELL-1 test set with previous work and unsupervised pre-training approach. Results with `8k' represents training data are downsampled
			 to 8kHz sample rate. Results with `16k' represents the sample rate of training data is 16kHz. Previous
			  work on AISHELL-1 are conducted with 16kHz sample rate speech data. Models with asterisk are our baseline results without pre-training data. Relative Error Reduction Rates
			   (RERR) are calculated as the percentage of error reduction compared to baseline.}
			\vspace{0.35cm}
			\begin{tabular}{llllllll}
				\toprule
			\multirow{2}{*}{Model}	&\multirow{2}{*}{Method}	&\multirow{2}{*}{Hours}	&\multicolumn{1}{l}{Unsupervised}	&\multicolumn{2}{c}{HKUST}					&\multicolumn{2}{c}{AISHELL-1}				\\
			&											&											&Pre-training Data							&CER/\%				&RERR/\%		 		&CER/\%					&RERR/\%			\\ \midrule
			TDNN-hybrid \cite{lfmmi}					&-					&-					&-							&23.7 						&-					&7.5 (16k)					&-					\\ 
			LSTM enc + LSTM dec \cite{DongZCX18}		&-						&-					&-							&29.4 					&-					&-						&-					\\
			Transformer \cite{karita2019comparative}	&-						&-					&-							&23.5 					&-					&6.7 (16k)					&-					\\ \midrule
			LSTM enc + LSTM dec*               			&- 						&-					&-							&28.8 		   			&- 					&12.9 (8k)				&-					\\
			LSTM enc + LSTM dec              			&APC					&$\sim$1500			&Open Mandarin				&27.8 		      		&3.5 				&11.4 (8k)				&11.6				\\ \midrule
			Transformer*		               			&- 						&					&-							&23.8 		   			&- 					&9.5 (8k)				&-					\\
			Transformer               					&MPC					&168				&HKUST						&23.3 		      		&2.1 				&-						&-				\\
			Transformer               					&MPC					&$\sim$1500			&Open Mandarin				&22.9 		      		&3.8 				&8.1 (8k)				&14.7				\\
			Transformer	 								&MPC			 		&5000				&Didi Callcenter			&21.7 	 				&8.8				&7.8 (8k)				&17.9				\\ 
			Transformer	 								&MPC			 		&10000				&Didi Callcenter			&\textbf{21.0} 		 	&\textbf{11.8}		&7.7 (8k)				&18.9				\\ 
			Transformer	 								&MPC			 		&10000				&Didi Dictation				&22.2 	 				&6.7				&\textbf{7.4} (8k)		&\textbf{22.1}		\\ 
			\bottomrule

			\end{tabular}
			\label{loss_HKUST}
		\end{table*}
		AISHELL-1 \cite{BuDNWZ17}, aidatatang\_200zh \cite{aidatatang}, MAGICDATA Mandarin Chinese Read Speech Corpus \cite{magicdata}, Free ST Chinese Mandarin Corpus (ST-CMDS) \cite{ST}
		 and Primewords Chinese Corpus Set 1 \cite{primewords}. 
		Detailed information of these corpora is provided in Table \ref{datasets}. Note the development set and test set of HKUST and AISHELL-1 were not included in pre-training.
		We name the combination of them Open Mandarin in the following experiments and it contains about 1500 hours of speech in total.
		To understand the impact of pre-training data size and speaking style on downstream task, our internal dataset Didi Dictation
		 and Didi Callcenter were also included. Didi Dictation contains approximately 10,000 hours of speech collected from our internal mobile
		  dictation application. Didi Callcenter also contains approximately 10,000 hours of speech collected from phone calls between our user
		   and customer service staff. All of our internal data are anonymized and eligible to be used for research purposes. 

		The fine-tuning experiments are conducted on HKUST and AISHELL-1. For HKUST, speed perturbation of 0.9, 1.0 and 1.1 was used
		 on training data and per-speaker normalization was applied on FBANK features. For AISHELL, speed perturbation of 0.9, 1.0 and 1.1 was also used
		 on training data. All speech data are downsampled to 8kHz because the sample rate for
		  both HKUST and Didi Callcenter is 8kHz. Note downsampling would usually hurt the accuracy of speech recognition system \cite{moreno1994sources,
		   schwarz2004towards} and results for AISHELL-1 is usually reported on the original 16kHz data. 
\begin{table}[ht]
	\centering			
	\caption{Details of open mandarin datasets and our internal datasets. ST-CMDS contains about 100 hours of speech data.}
	\vspace{0.35cm}
	\begin{tabular}{lll}
		\toprule
	Datasets		  		&Hours	 							&Speaking Style 				\\ \midrule 
	HKUST 					&168 								&Spontaneous 					\\ 
	AISHELL-1 				&178 								&Reading						\\
	aidatatang\_200zh 		&200 								&Reading						\\
	MAGICDATA 				&755 								&Reading 						\\
	ST-CMDS					&$\sim$100							&Reading						\\
	Primewords 				&100 								&Reading						\\ \midrule
	Didi Callcenter 		&10000 								&Spontaneous 					\\ 
	Didi Dictation 			&10000 								&Reading 						\\ \bottomrule
	\end{tabular}
	\label{datasets}
\end{table}
	\subsection{Experimental setups}
	\label{sec:model_structure}
		Most of our experiments are conducted on Transformer based models using MPC as pre-training method. To provide comparison with previous unsupervised
		 pre-training methods, we also conducted experiments on LSTM based, sequence-to-sequence with attention models using APC with $ num\_step\_ahead = 3 $.
		  Mandarin characters are used as modeling units in all our experiments \cite{ZhouDXX18, ZouJZYL18}.

		For LSTM based models, we follow the model structure in \cite{DongZCX18}, which consists of 4 unidirectional LSTM layers with 480 cells
		 in the encoder network and 1 layer LSTM with 320 cells in the decoder network. Layer normalization and residual connection were found to be
		  very important for LSTM based models because the goal of pre-training is to learn parameters that is invariant to the scaling and the offset
		   of the input. Without these structures, the network tends to wipe out the weights learned during pre-training and start from scratch
		    for the downstream task. Layer normalization and residual connection were added between each encoder LSTM layers.

		 For Transformer based models, we use similar model structure as paper \cite{karita2019comparative} ($ e = 12 $, $ d = 6 $, $ d_{model} = 256 $,
		   $ d_{ff} = 2048 $ and $ d_{head} = 4 $) for both HKUST and AISHELL-1. Downsampling is conducted between every three 
		    encoder Transformer blocks, resulting in an 8-fold downsample in total.

		For pre-training, both model was trained on 4 GPU with a total batch size of 256 for 500k steps. We used the
		 Adam optimizer \cite{KingmaB14} and varied learning rate with warmup schedule \cite{VaswaniSPUJGKP17} according to the formula: 
		\begin{equation}
			lrate = k * d_{model}^{0.5} * min(n^{-0.5}, n*warmup\_n^{-1.5}),
		\end{equation}
		where $n$ is the step number. $k = 0.5$ and $warmup\_n = 8000$ were chosen for all experiments. In the fine-tuning stage,
		 we used a total batch size of 128 with the same learning rate schedule and divide learning rate by 10 only once when validation
		  loss doesn't decrease for 5 epochs. Scheduled sampling with a sample rate of 0.1 is applied to help reduce the effect of 
		  exposure bias. We also used $1e^{-5}$ L2 normalization on network weights. As for decoding, the model with lowest loss on validation set 
		  was chosen. CTC-joint decoding is performed as proposed in \cite{Kim_2017} with CTC weight 0.3. 

	\subsection{Unsupervised pre-training}
	\label{sec:pre-training for HKUST and AISHELL-1}
		Three representative benchmarks on HKUST and AISHELL-1 were displayed here. The first is a traditional pipelined model optimized
		 with Lattice-Free Maximum Mutual Information (LF-MMI) objective. The second is an end-to-end LSTM based model with CTC-like objective.
		  The third one is a recently released Transformer based model that achieved very good results on multiple ASR benchmarks.

		As shown in Table \ref{loss_HKUST}, our baseline for HKUST matches the performance of previous work
		 for both LSTM and Transformer based model. 
		 %delete!
		 The baseline result for AISHELL-1 is worse than previous work but still in reasonable range  
		  because we performed downsample on speech.

		We first conducted experiments with APC and MPC using Open Mandarin as pre-training data. The work on APC \cite{Chung_2019} mainly presented 
		results with phone classification and speaker verification. Here we used it directly on the task of ASR and got 3.5\% relative error reduction 
		over a strong baseline. As shown in Table \ref{loss_HKUST}, the relative error reduction for MPC is similar to APC using the same pre-training data.

		In order to evaluate the performance improvements of the proposed method, we also conducted an experiment using only the HKUST training set to perform MPC
		 un-supervised pre-training and fine-tuning. The result listed in Table \ref{loss_HKUST} shows that 0.5\% absolutely CER reduction can be obtained, exceeding 
		 the best end-to-end model using only HKUST training set from 23.5\%\cite{karita2019comparative} to 23.3\%.

		To analyze the effect of pre-training data size, experiments were conducted with models pre-trained using different amounts of internal 
		data with the same speaking style. The relative CER reduction increase from 3.8\% to 8.8\% using 5000h of Didi Callcenter as
		 pre-training data. Adding another 5000h of Didi Callcenter further reduces CER by 11.8\% over the baseline. The same trend can be also observed for AISHELL-1.

		Speaking style has a strong impact on performance of ASR systems \cite{benzeghiba2007automatic, weintraub1996effect}.
		 To test whether speaking style of pre-training data would affect the performance of downstream tasks, further experiments 
		 with another pre-training model trained on 10000h Didi Dictation data is conducted.
		 Comparing relative error reduction for HKUST and AISHELL-1 with our internal data, models pre-trained with matching speaking style 
		 were found to brings more improvements for downstream tasks.
		 % In fact, the majority of Open Mandarin consists of reading type data and it helps more for AISHELL-1.
		 Interestingly, we noticed pre-training HKUST with 5000h of Didi Callcenter gets better performance than 10000h of Didi Dictation, which 
		  adds another proof for the importance of matching speaker style.

		To get a better understanding of MPC, we also conducted experiments with different pre-training steps
		 and compared unsupervised pre-training of MPC with supervised adaption. The following sections explain our findings in detail.

	\subsection{Effect of pre-training steps}
	\label{sec:pre-training steps}
\begin{figure}[h]
		\begin{minipage}[b]{1.0\linewidth}
			\centering
			\centerline{\includegraphics[clip, trim=2.5cm 8.8cm 3cm 9.5cm, width=0.93\linewidth]{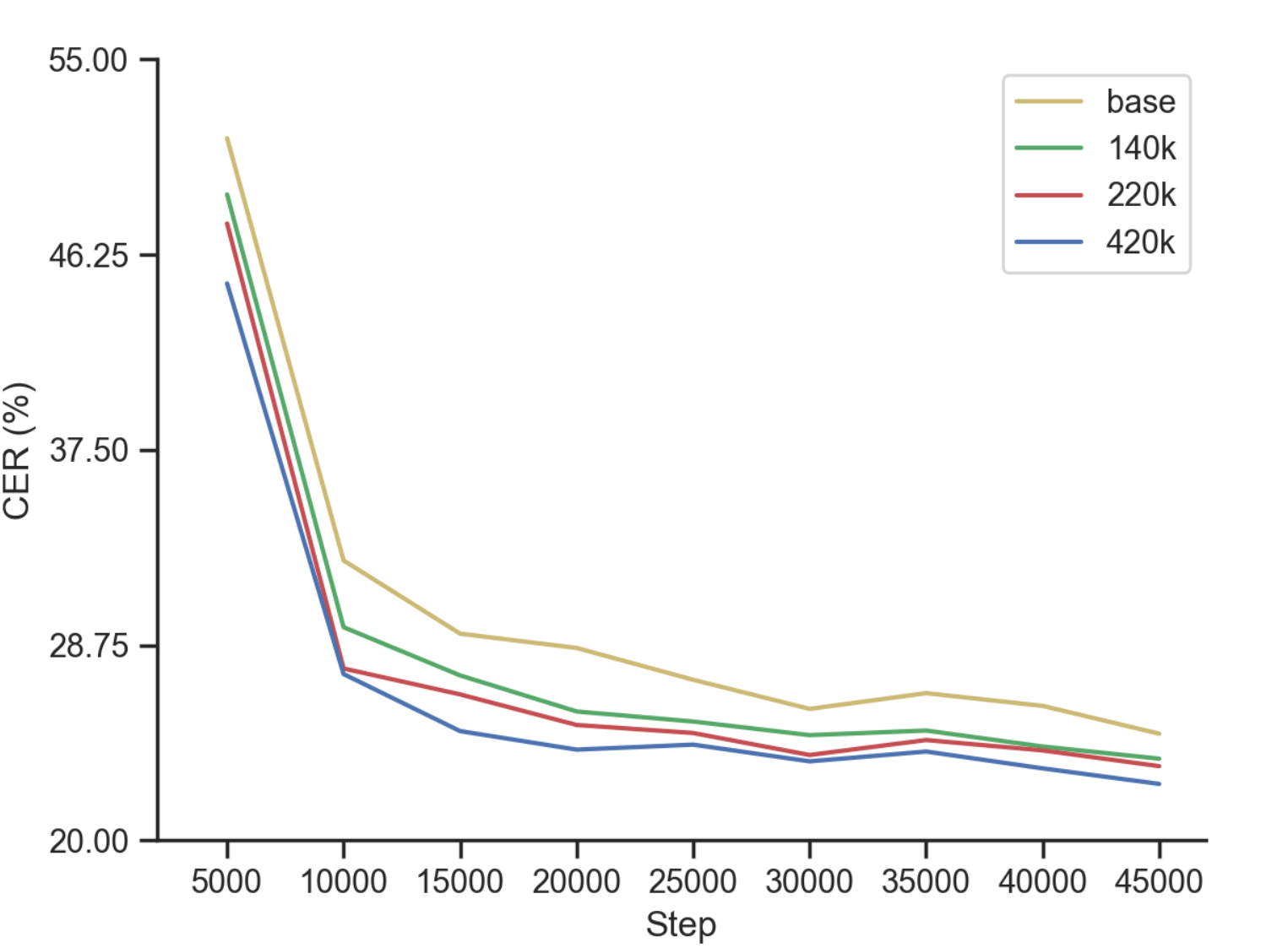}}
		\end{minipage}
	
		\caption{Convergence curve with different pre-training steps.}
		\label{fig:curve}
	\end{figure}

		It is known in the NLP community that more pre-training steps give a bigger performance boost for the 
		downstream task \cite{DBLP:conf/naacl/DevlinCLT19, liu2019roberta}. We also performed fine-tuning on HKUST with
		 pre-training model on Didi Callcenter at different steps with similar loss on validation set.
		  Fig. \ref{fig:curve} shows models pre-trained with more steps does give a slight
		   performance boost. The use of unsupervised pre-training also makes speech recognition
		    model converge a lot faster and more pre-training steps help in that direction.
	
	\subsection{Comparing with supervised adaption}
	\label{sec:comparsion}
		In this section, we present comparison between unsupervised pre-training and supervised adaption. For supervised adaption, 
		we first train a base model using both audio and transcript from Open Mandarin then fine-tune the model on HKUST and AISHELL-1.
		 As shown in Table
		  \ref{supervised_data}, the performance of supervised adaption is still better than our best unsupervised pre-training results on 
		  both HKUST and AISHELL-1 despite using less data. However, our proposed method does not need any transcribed data and it reduces
		   the cost to build a good speech recognition system significantly.

		\begin{table}[ht]
			\centering
			\caption{Comparison of Character Error Rates between unsupervised pre-training and supervised adaption on the test set of HKUST and AISHELL-1.}
			\vspace{0.35cm}
			\begin{tabular}{llcc}
				\toprule
			Model		        												& CER/\%			\\ \midrule 
			HKUST (Open Mandarin, Supervised Adaption)         					& 20.3			\\ 
			HKUST (Didi Callcenter, MPC Pre-training)     						& 21.0 			\\ \midrule
			AISHELL-1 (Open Mandarin, Supervised Adaption) 						& 7.0			\\
			AISHELL-1 (Didi Dictation, MPC Pre-training)       					& 7.4			\\ \bottomrule
			\end{tabular}
			\label{supervised_data}
		\end{table}

\section{Discussions and Conclusion}
\label{sec:conclusion}
	In this paper, we proposed Masked Predictive Coding (MPC), a method utilizing Masked-LM like structure for Transformer based
	 speech recognition models. Experimental results showed pre-training model with MPC is helpful to improve
	  performance on supervised speech recognition tasks. Experiments on HKUST
	  show that using the same training data, we can achieve CER 23.3\%,  exceeding the best end-to-end model by over 0.2\% absolute CER. 
	  With more pre-training data, we can further reduce the CER to 21.0\%, or
	   a 11.8\% relative CER reduction over baseline. Results of our work and APC seem to suggest language model-like
	   pre-training objectives can also help in speech tasks. We would like to explore the usefulness of 
	   other language model based pre-training objectives \cite{DaiYYCLS19, yang2019xlnet} in speech tasks.

	% Apart from speech recognition, Transformer based model has been integrated into many speech tasks like speech synthesis, 
	% speech emotion recognition and speaker verification \cite{karita2019comparative}.
	%  Our proposed objective has the potential to give performance boost to all these tasks. Our next step is to 
	%  test the effectiveness of our method on all those tasks.

	The experimental setup in this paper is very similar to the limited resource scenario for industrial applications.
	 Training set size of industrial applications is usually around hundreds to thousands of hours, but to achieve
	  superior performance, tens of thousands of hours is usually needed. On the other hand, when an 
	   applications goes online, it is very easy to collect lots of unsupervised data. Our experiments reveled 
	   the performance of MPC increases steadily with the increase of pre-training data, which makes it especially
		useful in limited resource scenarios. In the future, we will conduct experiments with ten times or even a
		 hundred times more unsupervised data to greatly reduce the gap between supervised adaption and
		  enable effective fine-tuning for industrial applications.
	
	In this work, speaking style in pre-training speech data was found to have a big impact on the performance of fine-tuning tasks and those having
	 similar speaking style to target dataset were found to be more helpful. Apart from speaking style, other speech variabilities like speaking environment,
	  speech scenarios and regional dialect \cite{benzeghiba2007automatic, qian2016very, li2018multi}
	  also have an impact on the performance of speech recognition system. Lots of recent advance in NLP is driven by pre-training
	   on texts with different style from different sources. Inspired by these advances, our next step is to perform experiments with 
	  unsupervised speech collected from different scenarios. We believe doing so would improve the performance of pre-training model and 
	  make it more robust.

\vfill\pagebreak

\bibliographystyle{IEEEbib}
\bibliography{wav2vecbib}

\begin{thebibliography}{10}

\bibitem{DBLP:conf/iccv/DoerschGE15}
C.~Doersch, A.~Gupta, and A.~Efros,
\newblock ``Unsupervised visual representation learning by context
  prediction,''
\newblock in {\em {ICCV}}, 2015, pp. 1422--1430.

\bibitem{DBLP:conf/naacl/DevlinCLT19}
J.~Devlin, M.~Chang, K.~Lee, and K.~Toutanova,
\newblock ``{BERT:} pre-training of deep bidirectional transformers for
  language understanding,''
\newblock in {\em {NAACL-HLT} {(1)}}, 2019, pp. 4171--4186.

\bibitem{DBLP:conf/naacl/EdunovBA19}
S.~Edunov, A.~Baevski, and M.~Auli,
\newblock ``Pre-trained language model representations for language
  generation,''
\newblock in {\em {NAACL-HLT} {(1)}}, 2019, pp. 4052--4059.

\bibitem{radford2018improving}
R.~Alec, N.~Karthik, S.~Tim, and S.~Ilya,
\newblock ``Improving language understanding with unsupervised learning,''
\newblock Tech. {R}ep., Technical report, OpenAI, 2018.

\bibitem{Schneider_2019}
S.~Steffen, B.~Alexei, C.~Ronan, and A.~Michael,
\newblock ``wav2vec: Unsupervised pre-training for speech recognition,''
\newblock {\em Interspeech 2019}, Sep 2019.

\bibitem{lian2018improving}
Z.~Lian, Y.~Li, J.Tao, and J.~Huang,
\newblock ``Improving speech emotion recognition via transformer-based
  predictive coding through transfer learning,''
\newblock {\em arXiv preprint arXiv:1811.07691}, 2018.

\bibitem{oord2018representation}
O.~Aaron van den, Y.~Li, and V.~Oriol,
\newblock ``Representation learning with contrastive predictive coding,''
\newblock {\em arXiv preprint arXiv:1807.03748}, 2018.

\bibitem{Ravanelli_2019}
R.~Mirco and B.~Yoshua,
\newblock ``Learning speaker representations with mutual information,''
\newblock {\em Interspeech 2019}, Sep 2019.

\bibitem{Pascual_2019}
P.~Santiago, R.~Mirco, S.~Joan, B.~Antonio, and et~al,
\newblock ``Learning problem-agnostic speech representations from multiple
  self-supervised tasks,''
\newblock {\em Interspeech 2019}, Sep 2019.

\bibitem{Chung_2019}
C.~Yu-An, H.~Wei-Ning, T.~Hao, and G.~James,
\newblock ``An unsupervised autoregressive model for speech representation
  learning,''
\newblock {\em Interspeech 2019}, Sep 2019.

\bibitem{DBLP:conf/icassp/DongWX19}
L.~Dong, F.~Wang, and B.~Xu,
\newblock ``Self-attention aligner: {A} latency-control end-to-end model for
  {ASR} using self-attention network and chunk-hopping,''
\newblock in {\em {ICASSP}}, 2019, pp. 5656--5660.

\bibitem{karita2019comparative}
K.~Shigeki, C.~Nanxin, H.~Tomoki, H.~Takaaki, and et~al,
\newblock ``A comparative study on transformer vs rnn in speech applications,''
\newblock {\em arXiv preprint arXiv:1909.06317}, 2019.

\bibitem{dong2018speech}
L.~Dong, S.~Xu, and B.~Xu,
\newblock ``Speech-transformer: a no-recurrence sequence-to-sequence model for
  speech recognition,''
\newblock in {\em ICASSP}. IEEE, 2018, pp. 5884--5888.

\bibitem{ZhouDXX18}
S.~Zhou, L.~Dong, S.~Xu, and B.~Xu,
\newblock ``A comparison of modeling units in sequence-to-sequence speech
  recognition with the transformer on mandarin chinese,''
\newblock in {\em {ICONIP} {(5)}}, 2018, vol. 11305, pp. 210--220.

\bibitem{radford2019language}
R.~Alec, J.~Wu, C.~Rewon, D.~Luan, and et~al,
\newblock ``Language models are unsupervised multitask learners,''
\newblock {\em OpenAI Blog}, vol. 1, no. 8, 2019.

\bibitem{PetersNIGCLZ18}
M.~Peters, M.~Neumann, M.~Iyyer, M.~Gardner, and et~al,
\newblock ``Deep contextualized word representations,''
\newblock in {\em {NAACL-HLT}}, 2018, pp. 2227--2237.

\bibitem{liu2019roberta}
Y.~Liu, O.~Myle, G.~Naman, J.~Du, and et~al,
\newblock ``Roberta: A robustly optimized bert pretraining approach,''
\newblock {\em arXiv preprint arXiv:1907.11692}, 2019.

\bibitem{ChanJLV16}
W.~Chan, N.~Jaitly, Q.~Le, and O.~Vinyals,
\newblock ``Listen, attend and spell: {A} neural network for large vocabulary
  conversational speech recognition,''
\newblock in {\em {ICASSP}}, 2016, pp. 4960--4964.

\bibitem{DBLP:conf/iscslp/LiuFYCHG06}
Y.~Liu, P.~Fung, Y.~Yang, C.~Cieri, and et~al,
\newblock ``{HKUST/MTS:} {A} very large scale mandarin telephone speech
  corpus,''
\newblock in {\em {ISCSLP}}, 2006, vol. 4274, pp. 724--735.

\bibitem{lfmmi}
D.~Povey, V.~Peddinti, D.~Galvez, P.~Ghahremani, and et~al,
\newblock ``Purely sequence-trained neural networks for {ASR} based on
  lattice-free {MMI},''
\newblock in {\em {INTERSPEECH}}, 2016, pp. 2751--2755.

\bibitem{DongZCX18}
L.~Dong, S.~Zhou, W.~Chen, and B.~Xu,
\newblock ``Extending recurrent neural aligner for streaming end-to-end speech
  recognition in mandarin,''
\newblock in {\em {INTERSPEECH}}, 2018, pp. 816--820.

\bibitem{BuDNWZ17}
H.~Bu, J.~Du, X.~Na, B.~Wu, and et~al,
\newblock ``{AISHELL-1:} an open-source mandarin speech corpus and a speech
  recognition baseline,''
\newblock in {\em {O-COCOSDA}}, 2017, pp. 1--5.

\bibitem{aidatatang}
Beijing DataTang Technology~Co{.}{,} Ltd,
\newblock ``aidatatang{\_}200zh{,} {a free Chinese Mandarin speech corpus},''
\newblock .

\bibitem{magicdata}
{Magic Data Technology Co{.}{,} Ltd},
\newblock ``{MAGICDATA Mandarin Chinese Read Speech Corpus},''
  {\url{http://www.imagicdatatech.com/index.php/home/dataopensource/data_info/id/101}},
  2019.

\bibitem{ST}
Surfingtech,
\newblock ``{ST}{-}{CMDS}{-}20170001{\_}1 {Free} {ST} {Chinese} {Mandarin}
  {Corpus},''
\newblock .

\bibitem{primewords}
{Primewords Information Technology Co., Ltd.},
\newblock ``{Primewords Chinese Corpus Set 1},'' 2018,
\newblock \url{https://www.primewords.cn}.

\bibitem{moreno1994sources}
M.~Pedro and S.~Richard,
\newblock ``Sources of degradation of speech recognition in the telephone
  network,''
\newblock in {\em ICASSP}, 1994, pp. I--109.

\bibitem{schwarz2004towards}
S.~Petr, M.~Pavel, and {\v{C}}.~Jan,
\newblock ``Towards lower error rates in phoneme recognition,''
\newblock in {\em TSD}, 2004, pp. 465--472.

\bibitem{ZouJZYL18}
W.~Zou, D.~Jiang, S.~Zhao, G.~Yang, and et~al,
\newblock ``Comparable study of modeling units for end-to-end mandarin speech
  recognition,''
\newblock in {\em {ISCSLP}}. 2018, pp. 369--373, {IEEE}.

\bibitem{KingmaB14}
D.~Kingma and J.~Ba,
\newblock ``Adam: {A} method for stochastic optimization,''
\newblock in {\em {ICLR}}, 2015.

\bibitem{VaswaniSPUJGKP17}
A.~Vaswani, N.~Shazeer, N.~Parmar, J.~Uszkoreit, and et~al,
\newblock ``Attention is all you need,''
\newblock in {\em {NIPS}}, 2017, pp. 5998--6008.

\bibitem{Kim_2017}
K.~Suyoun, H.~Takaaki, and W.~Shinji,
\newblock ``Joint ctc-attention based end-to-end speech recognition using
  multi-task learning,''
\newblock {\em ICASSP}, Mar 2017.

\bibitem{benzeghiba2007automatic}
B.~Mohamed, D.~Renato, D.~Olivier, D.~Stephane, and et~al,
\newblock ``Automatic speech recognition and speech variability: A review,''
\newblock {\em Speech communication}, vol. 49, no. 10-11, pp. 763--786, 2007.

\bibitem{weintraub1996effect}
W.~Mitch, T.~Kelsey, H.~Kate, and S.~Amy,
\newblock ``Effect of speaking style on lvcsr performance,''
\newblock in {\em Proc. ICSLP}, 1996, vol.~96, pp. 16--19.

\bibitem{DaiYYCLS19}
Z.~Dai, Z.~Yang, Y.~Yang, J.~Carbonell, and et~al,
\newblock ``Transformer-xl: Attentive language models beyond a fixed-length
  context,''
\newblock in {\em {ACL} {(1)}}, 2019, pp. 2978--2988.

\bibitem{yang2019xlnet}
Z.~Yang, Z.~Dai, Y.~Yang, J.~Carbonell, and et~al,
\newblock ``Xlnet: Generalized autoregressive pretraining for language
  understanding,'' 2019.

\bibitem{qian2016very}
Y.~Qian, M.~Bi, T.~Tan, and K.~Yu,
\newblock ``Very deep convolutional neural networks for noise robust speech
  recognition,''
\newblock {\em IEEE/ACM Transactions on Audio, Speech, and Language
  Processing}, vol. 24, no. 12, pp. 2263--2276, 2016.

\bibitem{li2018multi}
L.~Bo, S.~Tara N, S.~Khe Chai, B.~Michiel, and et~al,
\newblock ``Multi-dialect speech recognition with a single sequence-to-sequence
  model,''
\newblock in {\em ICASSP}, 2018, pp. 4749--4753.

\end{thebibliography}

\end{document}